# FPGA-based Acceleration of Neural Network for Image Classification using Vitis AI

Zhengdong Li[1], Frederick Ziyang Hong[2], C. Patrick Yue[3]

[1, 2, 3] Department of Electronic and Computer Engineering, The Hong Kong University of Science and Technology, Hong Kong S.A.R, CHINA

*Abstract*— In recent years, Convolutional Neural Networks (CNNs) have been widely adopted in computer vision. Complex CNN architecture running on CPU or GPU has either insufficient throughput or prohibitive power consumption. Hence, there is a need to have dedicated hardware to accelerate the computation workload to solve these limitations. In this paper, we accelerate a CNN for image classification with the CIFAR-10 dataset using Vitis-AI on Xilinx Zynq UltraScale+ MPSoC ZCU104 FPGA evaluation board. The work achieves 3.33-5.82x higher throughput and 3.39-6.30x higher energy efficiency than CPU and GPU baselines. It shows the potential to extract 2D features for downstream tasks, such as depth estimation and 3D reconstruction.

## I. INTRODUCTION

Nowadays, with the rapid evolution of deep learning, a lot of advanced CNN architectures have been proposed for various tasks, such as image classification, object detection, 3D reconstruction, etc. Current systems depending on CPU to run those algorithms are hard to meet the real-time data processing requirement. It also has low efficiency on power and performance. Thus, more and more neural networks are accelerated by GPUs to enhance their throughput. However, GPU consumes lots of power. Field Programmable Gate Array (FPGA) helps fill in the gap with its higher energy efficiency. Meanwhile, compared to ASIC, it is more flexible as well because the logic inside the chip is reconfigurable. Among multiple approaches of FPGA-based acceleration on neural networks, [1] made an implementation using Verilog. However, using hardware description language (HDL) to describe neural networks is very time-consuming without the help of third-party libraries, such as PyTorch or TensorFlow. One common alternative is to use High-Level Synthesis (HLS) to synthesize C/C++ functions into Register Transfer Level (RTL) codes. For example, [2, 3] leverage Vitis HLS to achieve the acceleration on FPGA. Yet, not all the layers of a CNN can be written in C/C++ in an efficient way, resulting in a longer development step.

Thus, in this paper, we leverage the advantage of Vitis-AI [13], introduced by Xilinx in 2019, an advanced development framework for AI inference on Xilinx hardware platforms that support both python and C/C++ coding. However, the evaluations of current research using Vitis-AI are not comprehensive enough. For instance, [4] leverages Vitis-AI to speed up neural networks for object detection, but the exact accuracy is not shown.

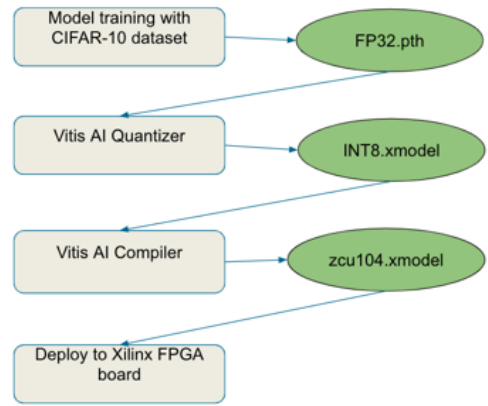

Fig. 1. The overall proposed working flow

Hence, in this paper, we evaluate the performance of the model running on an FPGA board with comprehensive metrics. The development pipeline is shown in Fig. 1. We first do pre-processing, training the network with the CIFAR-10 dataset in 32-bit floating point (FP32), then a quantization process to obtain an 8-bit integer (INT8) model, particularly in Xilinx model (.xmodel) format. After that, we compile the quantized model to include the instructions to be executed by the Deep Learning Processing Unit (DPU) on the target FPGA board. Finally we deploy the compiled model on ZCU104 [14] for testing. The design runs at 300MHz with dual B4096 DPU cores on ZCU104.

The key contributions of this paper are listed below:

- Implementation of an FPGA-based acceleration using Vitis-AI of CNN for a comprehensive comparison with multiple evaluation metrics to both CPU and GPU.

- The on board testing on ZCU104 manifest the proposed implementation reaches a higher throughput with a higher energy efficiency compared to CPU and GPU.

## II. VITIS-AI DESIGN FLOW

### A. Pre-processing

The neural network of this paper was inspired by CDRNet [5], a CNN for real-time 3D semantic meshing. We first extract the backbone of the CDRNet, which serves as the 2D feature extractor, including the first 35 Conv2d layers. We then strengthen the neural network with MnasNet [6] for image classification tasks up to 52 Conv2d layers to increase accuracy. Table I shows the network architecture details. The input of the model is an RGB image with a resolution of 32x32 from the CIFAR-10 dataset [7], training on 30 epochs, batch size 100, and learning rate 0.001 with 50000 training images while another 10000 images are used for testing. The procedure is done in Intel® Core™ i5-6600 to get the performance running on the CPU. We then repeat a similar approach for GPU, running on Nvidia RTX 3090 with CUDA version 11.4.

TABLE I. COMPARISON OF NETWORK ARCHITECTURE

| model | #layers (Conv2d) | $K$ | $S$ | $C_{in}$ | $C_{out}$ | #params |
|---|---|---|---|---|---|---|
| CDRNet (backbone part) | 35 | 3 or 5 | 1 or 2 | 3 | {80, 40, 24} | 321K |
| Ours | 52 | 1 or 3 or 5 | 1 or 2 | 3 | 1280 | 3120K |

### B. Quantization

After training the model in FP32, we quantize the model into INT8 using Vitis-AI Quantizer to reduce computational complexity, particularly in weights and activations layers like ReLU. It helps reduce the memory bandwidth from DPU and achieve a faster computational time. Since the Xilinx DPU family executes the model with the parameters in integer format, quantization is an essential process in the Vitis-AI design flow. Post Training Quantization (PTQ) is used in our case which is to calibrate and fine-tune a pre-trained model with unlabelled images, the testing dataset from CIFAR-10. Note that only batch size 1 is allowed before we export the .xmodel in Vitis-AI Quantizer. Thus, we first calibrate the quantization with the original batch size 100 and override it with batch size 1 to evaluate the quantized model.

### C. Compilation

The compilation is done using the Vitis-AI Compiler. It helps to compile the quantized model to include the optimized DPU instructions in .xmodel format which is a model to be stored in a MicroSD card executing by the ZCU104 later on. Various optimization techniques, like merging computation nodes are done to combine batch normalization with a preceding convolution. Others like data reuse and inherent parallelism are also done for efficient instruction scheduling. Before the deployment onboard, we have to ensure that the DPU subgraph number is equal to 1 for the output of the Vitis-AI Compiler to ensure that all layers have been compiled. Otherwise, we have to simplify the model by removing some of the unsupported layers. Note that we have to match the fingerprint of our ZCU104 evaluation board. This was done by storing the fingerprint in .json format during the compilation process, or the deployment on board cannot recognize the .xmodel.

### D. Deployment on board

Fig. 2 shows a demo of a on board testing for the compiled model running on ZCU104. We follow the default target reference design for the board image on ZCU104 before deployment on the FPGA board. The convolution architecture for parallelism can be configured, by selecting from B512, B800, B1024, B1152, B1600, B2304, B3136, and B4096. The number represents the peak operations per cycle. Hence, B4096 is used to maximize the performance for 4096 peak operations. Besides, we can choose at most 4 cores in the DPUCZDX8G IP architecture. However, more cores consume higher programmable logic (PL) resources and we observe that 3 or more cores of B4096 will result in implementation failure with insufficient resources after synthesis, particularly in BRAM. Hence, 2 cores of DPU are used and the estimated resource utilization is shown in Table II.

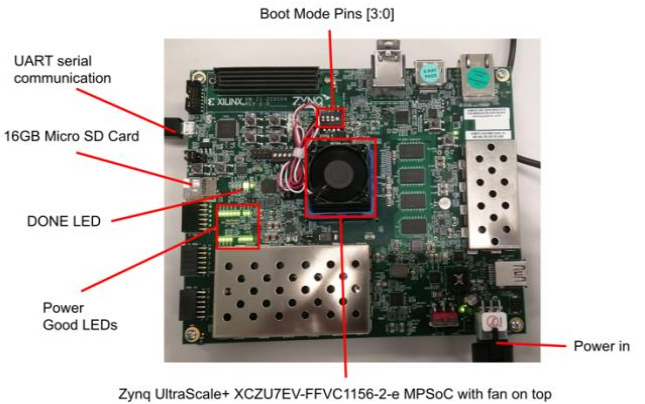

Fig. 2. On board testing in ZCU104: we use UART serial communication to the board via Minicom in Ubuntu 22.04 and Boot Mode Pins is set to 1110 for booting with a 16GB Micro SD Card where stores the board image, testing images, application codes running board and the .xmodel. The two rows of Power Good LEDs glow green show that the power system is working in normal condition and the system is ready when the DONE LED glows green.

TABLE II. RESOURCE UTILIZATION FROM VIVADO 2023.1

| DSP | BRAM | FF | LUT |
|---|---|---|---|
| 1420 (82.18%) | 210 (67.31%) | 198725 (43.13%) | 105845 (45.94%) |

### III. EVALUATION

Multiple threading is used for the comparison of throughout which is defined as the Frame Per Second (FPS) on board. The input images will be divided into subsets based on the number of threads, executing on DPU concurrently. [8] shows that there is an improvement in energy efficiency when increasing the number of computing threads to 4. In our experiment, FPS increases when the number of parallel threads increases up to 2. It is found that 584.11 FPS, 1021.45 FPS, and 920.81 FPS in 1-threading, 2-threading, and 3-threading respectively. We observe that 3-threadings or above will not further increase FPS in our case. One of the reasons is the limitation on memory bandwidth, meaning that the memory cannot handle the increased data transfer demands from the additional threads, resulting in reduced performance. It can only handle up to 2-threadings running simultaneously. Competing to access memory leads to potential bottlenecks.

The overall performance comparison of CPU, GPU, and FPGA is summarized in Table III. In terms of throughput, FPGA is 3.33-5.82x higher than CPU. In terms of energy efficiency, FPGA is also 3.39-6.30x higher than CPU. Besides, we also use Vitis Analyzer for profiling the hardware-level measurement, the DPU performance, which has 0.727 GOPS and 2041.91 MB/s in memory bandwidth.

TABLE III. OVERALL PERFORMANCE OF CPU VS GPU VS FPGA OF OUR NEURAL NETWORKS.

| Platform | CPU | GPU | FPGA |
|---|---|---|---|

|  |  |  | 1-thread | 2-threads |
|---|---|---|---|---|
| Vendor | Intel | Nvidia | Xilinx | |
| Type | I5-6600 | GeForce RTX 3090 | ZCU104 | |
| Technology (nm) | 14 | 8 | 16 | |
| Power (W) | 65 | 350 | 60 | |
| Frequency (GHz) | 3.3 | 1.4 | 0.3 | |
| Precision | FP32 | FP32 | INT8 | |
| Accuracy (%) | 68.62 | 68.61 | 58.76 | |
| Latency (s) | 56.99 | 44.78 | 17.12 | 9.79 |
| Throughput (FPS) | 175.47 | 223.31 | 584.11 | 1021.45 |
| Energy Efficiency (FPS/W) | 2.70 | 0.64 | 9.14 | 17.02 |
| **Throughput Comparison** | 1.0 | 1.27x | 3.33x | 5.82x |
| **Energy Efficiency Comparison** | 1.0 | 0.24x | 3.39x | 6.30x |

Furthermore, we run the backbone of the CDRNet described in Table I with the CIFAR-10 dataset and obtain the following performance in the same GPU: 62.77% accuracy, 24.51s latency, 408.00 FPS, and 1.17 FPS/W. Compared to our network running on an FPGA, we have 1.43-2.5x higher throughput and 7.81-14.55x higher energy efficiency. We believe that the acceleration of our 2D neural network can serve as preliminary work for the 2.5D depth map estimation and 3D model reconstruction in the future.

We also compare our performance with other relevant FPGA-based accelerations on CNN. Table IV shows that we have the highest throughput in terms of FPS. This helps to justify the significance of our work with diverse evaluation metrics.

TABLE IV.    COMPARISON WITH RELATED WORKS FOR IMAGE CLASSIFICATION TASK

|  | [9] | [10] | [11] | [12] | Ours |
|---|---|---|---|---|---|
| Publication, year | ICSICT, 2018 | ICSICT, 2018 | IEEE Access, 2023 | CCWC, 2022 | N.A. |
| FPGA Board | Xilinx ZC702 | Xilinx Zynq XC7Z100 | Xilinx ZCU104 | Xilinx Zynq-7000 | Xilinx ZCU104 |
| Dataset | CIFAR-10 | ImageNet | CIFAR-10 | CIFAR-10 | CIFAR-10 |
| Accuracy | 60~70% | N.A. | 30~85% | 66.6~92.5% | 58.76% |
| Throughput in FPS | N.A. | N.A. | 619.27 FPS | 68.8~441.3 FPS | 584.11~1021.45 FPS |
| Throughput in GOPS | 3.74 GOPS | 6.63 GOPS | N.A. | N.A. | 0.727 GOPS |

## IV. SUMMARY

In this paper, we present an FPGA-based acceleration on 2D CNN for image classification, achieving up to 5.82x and 6.30x higher in throughput and energy efficiency with 300MHz in dual B4096 DPU cores on ZCU104. The proposed workflow in Vitis-AI is useful and can serve as an important milestone to promote hardware acceleration of 3D model reconstruction for the whole neural network, such as CDRNet.

ACKNOWLEDGMENT

This work is in part supported by Bright Dream Robotics (BDR) and the HKUST-BDR Joint Research Institute Funding Scheme under Project HBJRIFTP-005 (OKT22EG06).